\documentclass[letterpaper, 10 pt, conference]{ieeeconf}  

\IEEEoverridecommandlockouts                              
\usepackage[utf8]{inputenc}
\usepackage[T1]{fontenc}
\usepackage{lipsum}  
\usepackage{siunitx}
\usepackage{graphicx}
\usepackage{hyperref}

\title{\LARGE \bf
Infrared and 3D skeleton feature fusion for RGB-D action recognition
}

\author{ \parbox{3 in}{\centering Alban Main de Boissiere\\
        Laboratoire de traitement de l’information en santé\\
        École de Technologie supérieure\\
        1100 Rue Notre-Dame W, Montréal, QC H3C 1K3\\
        {\tt\small alban.main-de-boissiere.1@ens.etsmtl.ca}}
        \hspace*{ 0.5 in}
        \parbox{3 in}{ \centering Rita Noumeir\\
        Laboratoire de traitement de l’information en santé\\
        École de Technologie supérieure\\
        1100 Rue Notre-Dame W, Montréal, QC H3C 1K3\\
        {\tt\small rita.noumeir@etsmtl.ca}}
}

\begin{document}

\maketitle
\thispagestyle{empty}
\pagestyle{empty}

\begin{abstract}

A challenge of skeleton-based action recognition is the difficulty to classify actions with similar motions and object-related actions. Visual clues from other streams help in that regard. RGB data are sensible to illumination conditions, thus unusable in the dark. To alleviate this issue and still benefit from a visual stream, we propose a modular network (FUSION) combining skeleton and infrared data. A 2D convolutional neural network (CNN) is used as a pose module to extract features from skeleton data. A 3D CNN is used as an infrared module to extract visual cues from videos. Both feature vectors are then concatenated and exploited conjointly using a multilayer perceptron (MLP). Skeleton data also condition the infrared videos, providing a crop around the performing subjects and thus virtually focusing the attention of the infrared module. Ablation studies show that using pre-trained networks on other large scale datasets as our modules and data augmentation yield considerable improvements on the action classification accuracy. The strong contribution of our cropping strategy is also demonstrated. We evaluate our method on the NTU RGB+D dataset, the largest dataset for human action recognition from depth cameras, and report state-of-the-art performances. 

\end{abstract}

\section{INTRODUCTION}
\PARstart{H}{uman} action recognition is an important computer vision field with applications ranging from video surveillance, robotics, to automated driving systems among others. It has been studied for decades but is still relevant due to its potential applications and recent rapid development \cite{wang2018rgb}. 

\begin{figure}[t]
  \centering
  \includegraphics[scale=0.6]{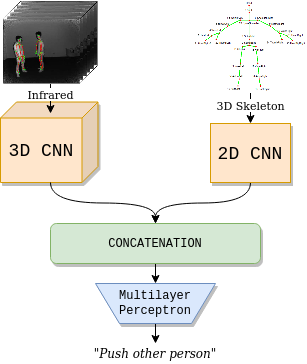}
  \caption{Our model uses a 2D CNN for pose data and a 3D CNN for IR sequences. Features from both modules are then concatenated and studied conjointly via an MLP. Training is done in end-to-end fashion.}
  \label{schematic_model}
\end{figure}

Consumer-grade depth cameras such as Intel RealSense \cite{keselman2017intel} and Microsoft Kinect \cite{zhang2012microsoft} coupled with advanced human pose estimation algorithms \cite{shotton2011real} have allowed 3D skeleton data to be obtained in real time. Key joints of the human body are extracted to a 3D space, providing a high-level representation of an action. Skeleton data are robust to surrounding  environment, illumination variations and may be generalized to various viewpoints \cite{aggarwal2014human}, \cite{han2017space}, \cite{presti20163d}, \cite{wang2018rgb}. Earlier works have indicated that key joints are powerful descriptors of human motion \cite{johansson1973visual}. The low dimensionality and high representation power make skeleton data a prime input for action recognition tasks. 

Opening the door for new action recognition algorithms, those are broadly categorized into RGB and 3D skeleton approaches. However, it has been demonstrated that visual and skeleton inputs can work in symbiosis \cite{rahmani2017learning}. Actions with similar body motion, such as writing versus typing on a keyboard, prove difficult to classify with skeleton data only. In this respect, skeleton data might benefit from the visual clues of RGB streams.

Depth cameras offer four different data streams: RGB, depth, infrared (IR) videos and 3D skeleton. To our knowledge, infrared videos from depth cameras have never been used as an input source for action recognition. We argue that the lack of large scale datasets proposing IR videos in addition to the other streams is in part responsible. Moreover, RGB and IR images are quite similar, the former offering a richer representation of a scene therefore making it a better candidate. However, IR is usable in the dark, which is viable for security applications when skeleton data are insufficient. The recent introduction of large scale datasets like NTU RGB+D \cite{shahroudy2016ntu} and PKU-MMD \cite{liu2017pku} containing IR videos motivates the evaluation of methods using this stream. Video understanding is a well-studied computer vision task. But modeling spatiotemporal features and long-term dependencies remains an issue. 
 
Another challenge in video action classification is the volume of information. To reduce the complexity of the videos, downscaling the frames is often employed but also comes with a decrease in the quality of the information. Moreover, discriminating clues may only occur in a small portion of the frames, becoming undetectable in the process. An alternative proposal is to focus on regions of interest. Visual attention models are capable of focusing on important cues and disregard other areas \cite{cho2015describing}, \cite{mnih2014recurrent}, \cite{sharma2015action}. 

In this work, we intend to address the difficulty of differentiating actions with similar motions with an additional visual stream insensible to illumination conditions. Furthermore, we evaluate the potential of IR videos as a standalone source. We propose a model fusing video and pose data (FUSION). Pose has a double purpose. It is used as an input stream in its own right and also conditions the IR sequences, providing a crop around the subjects, facilitating the classification. The general outline of the network is illustrated Fig. \ref{schematic_model}. 

The pose network is an 18-layer ResNet \cite{he2016deep} taking as input the entire skeleton sequence. The sequence is mapped to an RGB image which is then rescaled to fit the input size of the CNN. The IR network is a ResNet (2+1)D (R(2+1)D) \cite{tran2018closer} where a fixed number of random frames taken from evenly spaced subsequences are used as inputs. The features of each module are then fused using a concatenation scheme before proposing a final classification with a multilayer perceptron (MLP). 

Our main contributions are as follows :

\begin{itemize}
    \item We demonstrate the importance of IR streams from depth cameras for human action recognition.
    \item We propose a fusion network taking skeleton and IR sequences as inputs, which has never been attempted before. 
    \item We perform extensive ablation studies. We isolate different modules of our model and study their individual representation power. We also evaluate the importance of data augmentation, transfer learning, 2D-skeleton conditioned IR sequences and IR sequence length on accuracy score.
    \item We achieve state-of-the-art results compared to methods using different streams.
\end{itemize}{}

Codes, documentation and supplementary materials can be found on the project page. \footnote{\href{https://github.com/adeboissiere/FUSION-human-action-recognition}{Link to project page}}

\section{Related Work}

\subsection{Skeleton-based approaches}

Human action recognition has received a lot of attention due to its high-level representation and powerful discriminating nature. Traditional approaches focus on handcrafted features \cite{hussein2013human}, \cite{vemulapalli2014human},  \cite{wang2012mining}. These could be the dynamics of joint motion, covariance matrix of joint trajectories \cite{hussein2013human} or the representation of joints in a Lie group \cite{vemulapalli2014human}. Design choices prove challenging and result in suboptimal results. Recent deep-learning methods report improved accuracy. There exists three main frameworks: sequence-based models, image-based models and graph-based models. 

Sequence models exploit skeleton data as time series of key joints which are then fed to recurrent neural networks (RNN) \cite{du2015hierarchical}, \cite{lee2017ensemble}, \cite{liu2016spatio}, \cite{shahroudy2016ntu}, \cite{song2017end}, \cite{wang2017modeling}, \cite{zhang2017view}. Part-aware long short-term memory (LSTM) RNN \cite{shahroudy2016ntu} uses different memory cells for different regions of the body, then fuses them for the final classification. Similarly in \cite{du2015hierarchical}, a bidirectional RNN studies separate body parts individually in earlier levels and conjointly deeper on. In an effort to model simultaneously time and spatial dependencies, Liu \textit{et al.} propose a 2D recurrent model \cite{liu2016spatio}. Recurrent models are now part of the early deep learning efforts for skeleton-based action recognition. Vastly improving upon the results of the traditional methods, they remain insufficient. The sequence length has to be fixed during training which is not ideal and requires a sampling strategy. Moreover, sequence models tend to be much slower than their image-based counterpart.

Image models represent skeleton data as 2D images which are then used as inputs for convolutional neural networks (CNN) \cite{du2015skeleton}, \cite{ke2017new}, \cite{kim2017interpretable}, \cite{li2017skeleton}, \cite{liu2017enhanced}, \cite{wang2016action}. An intuitive method is to assign the $x$, $y$ and $z$ coordinates of a skeleton sequence to the channels of an RGB image \cite{du2015skeleton}, \cite{li2017skeleton}. Each joint corresponds to a row and each frame to a column, or inversely. Pixel intensity is then normalized between 0 and 255 based on maximal coordinates value of the dataset \cite{du2015skeleton} or sequence \cite{li2017skeleton}. Other works utilize the relative coordinates between joints to generate multiple images \cite{ke2017new}. Wang \textit{et al.} project the 3D coordinates on orthogonal 2D planes and encode the trajectories into a hue, saturation, value (HSV) space \cite{wang2016action}. A pre-trained model over ImageNet \cite{deng2009imagenet} is leveraged. A similar approach is used in \cite{hou2016skeleton}. More recent works focus on view-invariant transformations \cite{ke2017skeletonnet}, \cite{liu2017enhanced} or networks \cite{zhang2019view} with improved results. In \cite{kim2017interpretable}, a temporal convolutional network is deployed with interpretability of the results a major objective. CNNs are able to learn from entire sequences rather than sampled frames. The image generated from the skeleton sequence is resized to accommodate the fixed input shape of the CNN. This means an entire sequence can be used at once, which is an advantage compared to recurrent methods. 

Graph neural networks have received a lot of attention as of late due to their effective representation of skeleton data \cite{xu2018powerful}. There exists two main graph model architectures: graph neural networks (GNN), which combine graph and recurrent networks, and graph convolutional networks (GCN), which aim to generalize traditional convolutional networks. Of this architecture derives two types of GCNs: spectral and spatial. Spatial GCNs leverage the convolution operator for each node using its nearest neighbors \cite{simonovsky2017dynamic}. Yan \textit{et al.} \cite{yan2018spatial} make best of the graph representation to learn both spatial and temporal features. Li \textit{et al.} generalize the graph representation to actional and structural links \cite{li2019actional}. In \cite{si2019attention}, a temporal attention mechanism is adopted to enhance the classification while exploring the co-occurrence relationship between spatial and temporal domains. In \cite{shi2019two}, length and direction of bones are used in addition to joint coordinates while adapting the topology of the graph. Shi \textit{et al.} represent skeleton data as a directed acyclic graph based on kinematic dependencies of joints and bones \cite{shi2019skeleton}. GCNs report the current state-of-the-art results on benchmark datasets. However, carefully designed CNNs show comparable results \cite{zhang2019view}. Also, CNNs can be pre-trained on other large scale datasets which actually improves the performances of image-based skeleton action recognition models \cite{zhang2019view}. To our knowledge, an ImageNet \cite{deng2009imagenet} style transfer learning is impractical for GCNs.  

\subsection{RGB-based video classification} \label{review_rgb_methods}

\begin{figure}[t]
  \centering
  \includegraphics[scale=0.5]{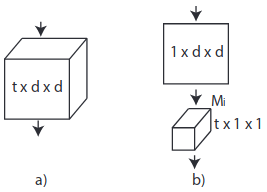}
  \caption{a) A standard 3D convolution operator. b) A factorized (2+1)D convolution operation with an additional non-linear activation function in between. Illustration courtesy of \cite{tran2018closer}.}
  \label{3Dconv}
\end{figure}

Traditional approaches focus on handcrafted features in the form of spatiotemporal interest points. Among those, improved Dense Trajectories (iDT) \cite{wang2013action}, which uses estimated camera movements for feature correction, is considered the state of the art. After the widespread use of deep learning on single images, many attempts have been made to propose benchmarks for video classification.

Soon after \cite{wang2013action}, two breakthrough papers \cite{karpathy2014large}, \cite{simonyan2014two} would form the backbone of future efforts. In \cite{karpathy2014large}, Karpathy \textit{et al.} explore different ways of fusing temporal information using pre-trained 2D CNNs. In \cite{simonyan2014two}, handcrafted features, in the form of optical flow, are used symbiotically with the raw video. Two parallel networks compute spatial and temporal features. A few drawbacks include the inability to effectively capture long-range temporal information and the heavy calculations required to compute optical flow. 

Later research propositions fall into five frameworks :
\begin{itemize}
    \item 2D CNN followed by RNN network \cite{donahue2015long}
    \item 3D CNN \cite{tran2015learning}, \cite{yao2015describing}
    \item Two-Stream 2D CNN \cite{feichtenhofer2016convolutional}
    \item 3D-Fused Two-Stream \cite{feichtenhofer2016convolutional}
    \item Two-Stream 3D CNN \cite{carreira2017quo}
\end{itemize}

Heavy networks and computations of handcrafted features as well as the absence of a benchmark for long-term temporal features remain an issue. In \cite{tran2018closer}, Tran \textit{et al.} explore different forms of spatiotemporal convolutions and their impact on video understanding. A (2+1)D convolution block separating spatial and temporal filters allows for a greater non-linearity compared to a standard 3D block with an equivalent number of parameters, as illustrated Fig. \ref{3Dconv}. Separating convolutions yields state-of-the-art results on benchmark datasets such as Sports-1M \cite{karpathy2014large}, Kinetics \cite{carreira2017quo}, UCF101 \cite{soomro2012ucf101} and HMDB51 \cite{kuehne2011hmdb}. 

\subsection{Mixed inputs action recognition}

Depth cameras provide different streams, or in other words, different representations of a same action. Some works have attempted to improve classification by combining streams. It can be argued that skeleton-based approaches prove most effective at discriminating actions with broad movements. However for actions involving similar joint positions and trajectories, such as reading vs. playing on a phone, skeleton-based models do not perform as well. Visual streams can provide important cues such as the type of object held. RGB and depth streams have been studied extensively. However, to our knowledge, we are the first to use IR data from depth cameras for action recognition.

In \cite{hu2018deep}, \cite{shahroudy2017deep}, \cite{wang2018cooperative} the complementary role of RGB and depth is demonstrated. In \cite{zolfaghari2017chained}, pose, motion and raw RGB images are inputted in 3 parallel 3D CNNs. Although visual information greatly improves upon the pose baseline, results are comparable with the then state-of-the-art methods using only skeleton data. In \cite{rahmani2017learning}, human-object interactions are modeled using both skeleton and depth data. An end-to-end network is proposed to learn view-invariant representations of skeleton data and held objects. Once again, visual information increases the accuracy but the results do not justify the complexity of a fusion approach compared to the other skeleton-only approaches of the time. The same year, Baradel \textit{et al.} use RGB and skeleton data conjointly in a pertinent way \cite{baradel2017pose}. Pose information is used as an input but also conditions the RGB stream. The 3D skeleton data are projected onto the RGB sequences to effectively extract crops around the hands of the subject, serving as another input. The RGB stream thus provides important clues about an object held and inter-subject interactions, significantly improving the results. This work shows that not all body parts need to be focused on, unlike the approach in \cite{rahmani2017learning}. But this requires as many streams as there are hands, which is memory inefficient. Furthermore, when the hands are close together, the information provided may be redundant. 

We propose a similar approach to \cite{baradel2017pose} and \cite{rahmani2017learning} in which the 3D skeleton data provide a crop around the subjects, alleviating the need for a spatial attention mechanism. A single crop is necessary, even when multiple subjects are interacting, which relaxes the memory needs.

\section{Proposed Model}

We design a deep neural network using skeleton and IR data, called "Full Use of Infrared and Skeleton in Optimized Network" (FUSION). The network consists of two parallel modules and an MLP. One module interprets skeleton data, the other IR videos. The features extracted from each individual stream are then fused using a concatenation scheme. The MLP is used as the final module and outputs a probability density. The network is trained in end-to-end fashion by optimizing the classification score. 

We note a skeleton sequence $S = \{S_{j,t,k}\}$ where $j$ denotes a joint index, $t$ a frame index and $k$ a coordinate axis ($X$, $Y$ and $Z$). We note $I = \{I_{t}\}$ a sampled IR sequence, as detailed section \ref{ir_sub_sample}, where $t$ is taken between $\{1, .., T\}$, with $T$ the number of sampled frames.

In the following sections, we present the individual modules of our FUSION model: a 2D CNN as the pose module, a 3D CNN as the IR module and an MLP as the stream fusion module. 

\subsection{Pose module}

\begin{figure}[t]
  \centering
  \includegraphics[width=\columnwidth]{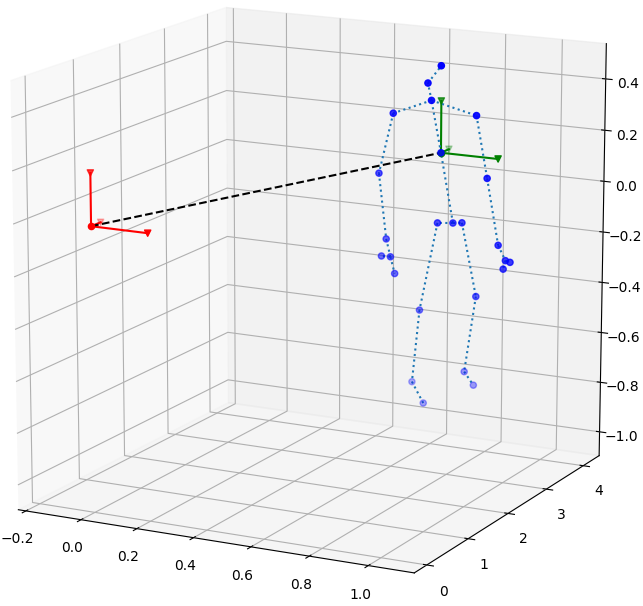}
  \caption{In red the coordinate system of the camera, in green the new coordinate system corresponding to the middle of the spine of the main subject for the first frame of the sequence, in blue the skeleton of the main subject, in black the translation vector.}
  \label{prior_norm_step}
\end{figure}

A skeleton sequence requires careful treatment for optimal results. First, a skeleton sequence is normalized to be position invariant, meaning the distance between the subject and the camera is accounted for. The sequence is then transcribed to an RGB image, with multiple subjects interactions in mind. The handcrafted RGB image is then fed to a 2D CNN. 

\subsubsection{Prior normalization step}

Each skeleton sequence is normalized by translating the global coordinate system of the camera to a local coordinate system corresponding to a key joint of the main subject. We choose the middle of the spine as the new origin. This is illustrated Fig. \ref{prior_norm_step}.

We adopt a sequence-wise normalization. In other words, the translation vector is computed for the first frame and applied to each subsequent frame, meaning the subject may move away from the new local coordinate system, as follows:

\begin{equation}
S' = S_{:, :, :} - S_{1, 0, :}.
\end{equation}

Where $S'$ is the normalized skeleton sequence, $j=1$ corresponds to the middle of the spine for the Kinect 2 skeleton \cite{zhang2012microsoft}. The "$:$" notation signifies that all values are considered across this dimension.

\subsubsection{Skeleton data to skeleton 2D maps}

A skeleton sequence is mapped to an image similar to \cite{du2015skeleton}, a skeleton map. Each coordinate axis, $X$, $Y$ and $Z$, is attributed to each channel of an RGB image. Each key joint corresponds to a row while the columns represent the different frames. 

We apply a dataset-wise normalization \cite{du2015skeleton}. We note $c_{min}$ and $c_{max}$ the minimal and maximal values of the coordinates after the normalization step for the entire dataset. The pixels of the skeleton map are recalculated using a min-max strategy in the $[0, 1]$ range, as follows: 

\begin{equation}
M = \frac{S' - c_{min}}{c_{max} - c_{min}}.
\end{equation}

Where $M = \{M_{j,t,k}\}$ is the normalized skeleton map with $k$ both the coordinate axis and the image channel. 

To accommodate for the fixed input size of the 2D CNN, the skeleton map is resized to a standard size.

\subsubsection{Multi subject strategy}

\begin{figure}[t]
  \centering
  \includegraphics[width=0.8\columnwidth]{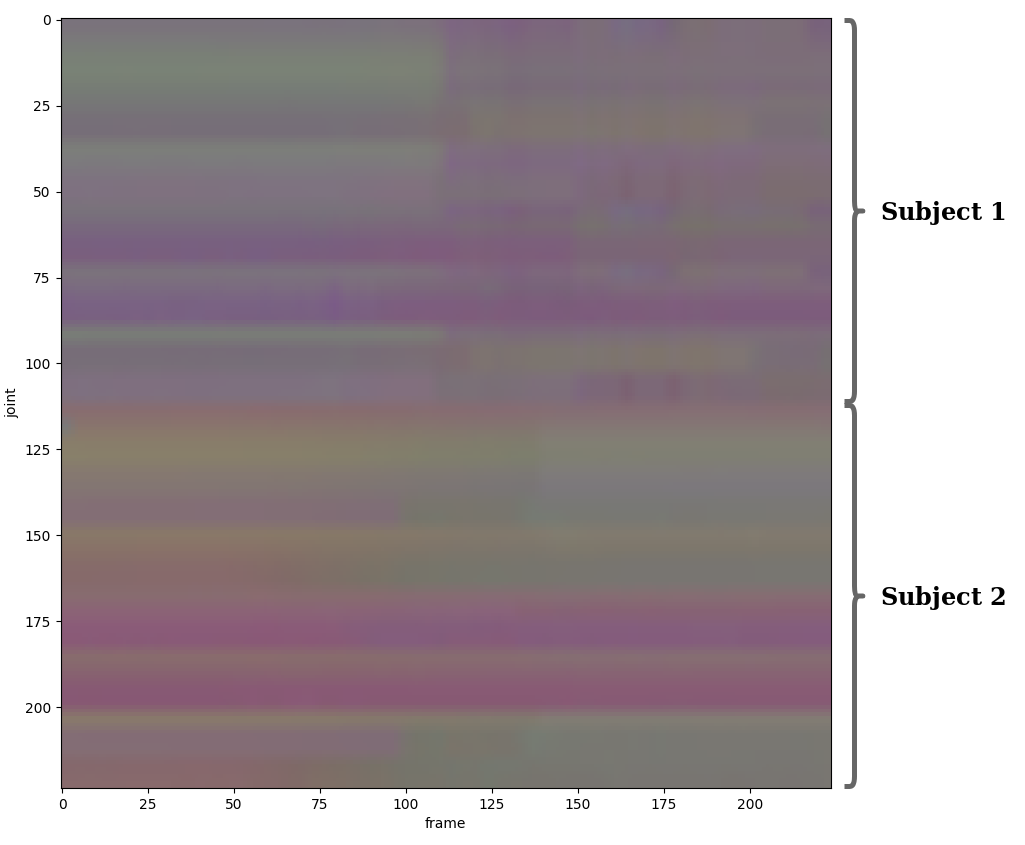}
  \caption{Skeleton map of two subjects. The joints of the two subjects are concatenated across a dimension, then stacked over time. The created image is reshaped to the fixed CNN input size.}
  \label{skeleton_map}
\end{figure}

Our network is scalable to multiple subjects. We concatenate the different skeleton maps across the joint dimension. With $J$ being the total number of joints, the first $J$ rows correspond to the first subject, the subsequent $J$ rows to subject 2, etc. We limit the number of subjects to two, corresponding to the maximum of the NTU RGB+D dataset \cite{shahroudy2016ntu}. Nonetheless, this method may be generalized to a greater number of subjects. Should the skeleton sequence comprise only of one subject, the $J$ rows of the second subject are set to zero. 

In case of multiple subjects, the coordinates of the latter are translated to the local coordinate system of the main subject (Fig. \ref{skeleton_map}).

The advantages of our method are manifold. Firstly, this alleviates the need for individual networks for different subjects. Secondly, this representation allows for a second subject to still intervene if its skeleton is detected after the first frame. Thirdly, the distance information is kept as each subject coordinates are translated to the local coordinate system of the first subject. Lastly, the skeleton map is resized to a standard size to accommodate for the fixed input size of the pose module. This implies that the network is able to learn from raw sequences of different sizes.

\subsubsection{CNN used}

The transformed skeleton map is used as input. We use an existing CNN with pre-trained weights on ImageNet as we find this ameliorates the classification score even when the images are handcrafted. We choose an 18-layer ResNet \cite{he2016deep} for its compromise between accuracy and speed. 

We extract a pose feature vector $s$ from the skeleton map $M$ with the pose module $f_S$ with parameters $\theta_S$ (\ref{eq:pose_module}). Here, and for the rest of the paper, subscripts of modules and parameters refer to a module, not an index.

\begin{equation} \label{eq:pose_module}
s = f_S(M|\theta_S)
\end{equation}

\subsection{IR module}

The action performed by a subject is only a small region inside the frames of an IR sequence. The 2D skeleton data are used to capture the region of interest and virtually focus the attention of the network, with multiple potential subjects in mind. Because the IR module requires a video input with a fixed number of frames, a subsampling strategy is deployed. A 3D CNN is used to exploit the IR data.  

\subsubsection{Cropping strategy} \label{cropping_strategy}

\begin{figure}[t]
  \centering
  \includegraphics[width=0.8\columnwidth]{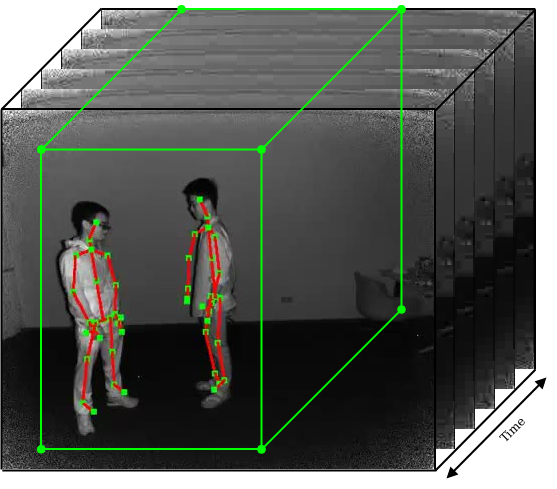}
  \caption{A fixed bounding box across the entire sequence is generated using the 2D skeleton information. The new sequence focuses attention on the subject rather than the background which provides little to no context. The images are taken from the NTU RGB+D dataset \cite{shahroudy2016ntu}.}
  \label{parallel_ir}
\end{figure}

Traditionally, 3D CNNs require a lot of parameters to account for the complex task of video understanding. Thus, the frames are heavily downscaled to reduce memory needs. In the process, discriminating information may be lost. In an action video of daily activities, the background provides little to no context. We would like our model to only focus on the subject as this is where the action happens. We argue that a crop around the subject provides ample cues about the action performed. Depth information, coupled with pose estimation algorithms, provides a turnkey solution for human detection. We propose a cropping strategy, shown Fig. \ref{parallel_ir} by a green parallelepiped, to virtually force the model to focus on the subject.

Given a 3D skeleton sequence projected on the 2D frames of the IR stream, we extract the maximal and minimal pixel positions across all joints and frames. This creates a fixed bounding box capturing the subject on the spatial and temporal domains. We empirically choose a 20 pixels offset to account for potential skeleton inaccuracy. The IR stream is padded with zeros should the box coordinates with the offset exceed the IR frame range. 

The advantage of our method is as follows. Providing a crop around the region of interest reduces the size of the frames without decreasing the quality. The downscaling factor is thus less important and preserves a better aspect of the image. Furthermore, it alleviates the need for an attention mechanism as the cropping strategy may be seen as a hard attention scheme in itself. Also, the network does not have to learn information from the background, which is noise in our case, as it is reduced to a minimum.

\subsubsection{Multi subject strategy}

The cropping strategy can be generalized to multiple subjects. The bounding box is enlarged to account for the other subjects. We take the maximal and minimal values across all joints, frames and subjects.

For a given sequence, the bounding box is immobile regardless of the number of subjects. This allows keeping camera dynamics. We do not want to add confusion to the sequence by adding a virtual movement of the camera with a mobile bounding box. 

\subsubsection{Sampling strategy} \label{ir_sub_sample}

\begin{figure}[t]
  \centering
  \includegraphics[width=\columnwidth]{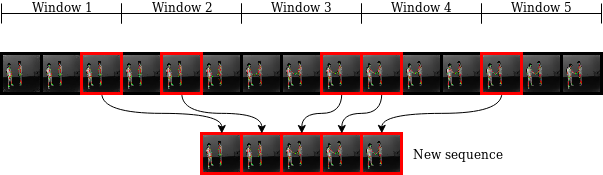}
  \caption{Each IR sequence is divided into a fixed number of windows of equal size. For each subdivision, a random frame is sampled. The concatenation of those frames is the input for the IR module.}
  \label{sub_windows}
\end{figure}

Contrary to the pose network, a given IR sequence is not treated in its entirety. A 3D CNN requires a sequence with a fixed number of frames $T$. Choices must be made regarding the value of $T$ and the sampling strategy. A potential approach would be to take adjacent frames in a sequence. But the subsequence might not be enough to correctly capture the essence of the action. Instead, we propose a scheme where the raw sequence is divided into $T$ windows of equal duration similar to \cite{liu2016spatio}, as illustrated Fig. \ref{sub_windows}. A random frame is taken from each window. A new sequence is created of length $T$. This is a form a data augmentation as a raw sequence may yield different results.

\begin{figure*}[t]
  \centering
  \includegraphics[width=\textwidth]{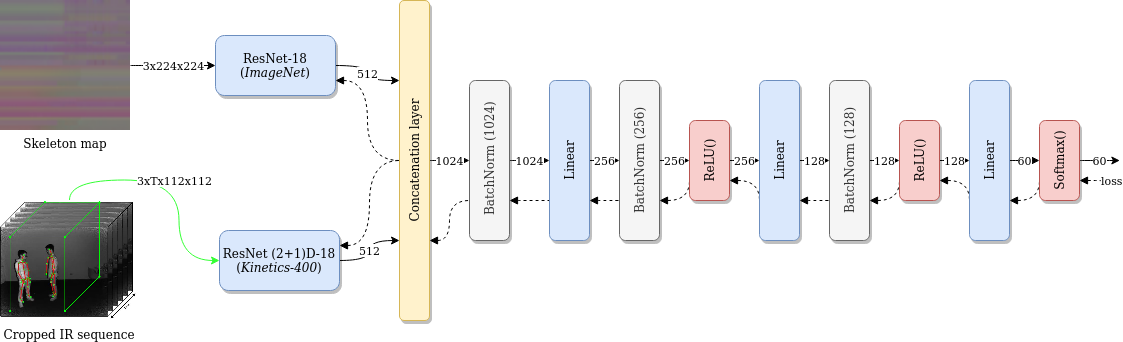}
  \caption{The full detailed model. The pose and IR modules output separate feature vectors. The two are concatenated and a final MLP outputs a class probability distribution. The pose network is a pre-trained \textit{ResNet-18}. The IR network is a pre-trained \textit{R(2+1)D-18} network.}
  \label{full_model}
\end{figure*}

\subsubsection{3D CNN used}

The new sampled sequences are used as inputs for the 3D CNN. We use an 18-layer deep R(2+1)D network \cite{tran2018closer} pre-trained on Kinetics-400 \cite{carreira2017quo}. R(2+1)D is an elegant network which revisits 3D convolutions. Tran \textit{et al.} showed factoring spatial and temporal convolutions yields state-of-the-art results on benchmark RGB action recognition datasets. Separating spatial and temporal convolutions with a nonlinear activation function in between allows for a more complex function representation with the same number of parameters.

We extract a stream feature vector $i$ from the sampled IR sequence $I$ with the IR module $f_{IR}$ with parameters $\theta_{IR}$, as follows:

\begin{equation}
i = f_{IR}(I|\theta_{IR}).
\end{equation}

\subsection{Stream fusion}

Both pose and IR modules output their own feature vectors. An MLP serves as the final module and returns a probability distribution for each action class in a dataset. 

Features of both streams are fused using a concatenation scheme. The MLP consists of three layers with batch normalization \cite{ioffe2015batch} before computation. The \textit{ReLU} activation function is used for all neurons. Lastly, a \textit{softmax} activation function is deployed to normalize the last layer's output into a probability distribution. 

The class probability distribution $y$ is outputted by the MLP $f_{MLP}$  with parameters $\theta_{MLP}$ (\ref{eq:MLP}). Inputs $i$ and $s$ correspond to the feature vectors computed by the pose and IR modules.

\begin{equation} \label{eq:MLP}
y = f_{MLP}(i, s|\theta_{MLP})
\end{equation}

We tried a scheme where the pose and IR modules of our network would emit their own prediction. We would then average the predictions on a logits level with learned weights during the backpropagation step. However, this would lead to the network's final classification to be attributed solely to one module or the other. Instead, we believe that an MLP allows for the features of the different streams to be interpreted conjointly.

\section{Network Architecture}
\subsection{Architecture}

\subsubsection{Pose module}

The pose network is an 18-layer deep ResNet \cite{he2016deep}. The network takes as input a tensor of dimensions 3x224x224, where 3 corresponds to the RGB channels and 224 to the height and width of the image. The output, $s$, is a 1D vector of 512 features.

\subsubsection{IR module}

The IR network is an 18-layer deep R(2+1)D \cite{tran2018closer}. It takes as input a video of dimensions 3xTx112x112, where 3 corresponds to the RGB channels, $T$ to the length of the sequence and 112 to the height and width of the image. The output, $i$, is a 1D vector of 512 features. 

To be able to leverage the pre-trained R(2+1)D CNN, which is originally trained on RGB images, the IR frames, which are single-channel grayscale images, are duplicated.

\subsubsection{Classification module}

The classification module is an MLP network with three layers. The first layer expects a vector of 1024 features and comprises 256 units. The second layer consists of 128 units. The last layer has as many units as there are different action classes in a dataset. Finally, the \textit{softmax} function is used to normalize the predictions to a probability distribution. Batch normalization is applied before the layers. A dropout scheme has been tested in place of batch normalization but was not found to be superior. The \textit{ReLU} activation function is used for all layers except the last.

The entire network is detailed Fig. \ref{full_model}.

\subsection{Data augmentation}

To prevent overfitting and reinforce the generalization capabilities of our model, we perform data augmentation during training. 

The skeleton sequences have limited viewpoints but their representation makes them excellent candidates for augmentation through geometric transformations. The skeleton sequences are enhanced by performing a random rotation around the $X$, $Y$ and $Z$ axis. For each sequence during training, we apply a random rotation between \ang{-20} and \ang{20} on each axis. 

We approach IR data augmentation with the following scheme. For each sequence during training, we perform a horizontal mirroring transformation on the frames with a 50\% chance probability. The two streams are augmented independently. 

\subsection{Training}

The network is trained in end-to-end fashion by minimizing cross-entropy loss, meaning all the modules of our network are trained together. The pose network is pre-trained on the ImageNet dataset \cite{deng2009imagenet}. The IR network is pre-trained on the Kinetics-400 dataset \cite{carreira2017quo}.

\section{Experiments}

We evaluate the performances of our proposed model on the NTU RGB+D dataset, the largest benchmark to date \cite{shahroudy2016ntu}. We also perform extensive ablation studies to understand the individual contributions of our modules.

\subsection{NTU RGB+D dataset}

The NTU RGB+D dataset is the largest human action recognition dataset to date captured with a Microsoft Kinect V2 \cite{zhang2012microsoft}. To our knowledge, it is also the only one including the IR sequences. It contains 60 different classes ranging from daily to health-related actions spread across 56,880 clips and 40 subjects. It includes 80 different views. An action may require up to two subjects. The various setups, views, orientations, result in a great diversity which makes NTU RGB+D a challenging dataset. 

There are two benchmark evaluations for this dataset: Cross-Subject (CS) and Cross-View (CV). The former splits the 40 subjects into training and testing groups. The latter uses the samples acquired from cameras 2 and 3 for training while the samples from camera 1 are used for testing. 

\subsection{Experimental settings}

\begin{table}[t]
\caption{Results of the pose module on NTU RGB+D dataset (accuracy in \%)}
\label{skeleton_block_results}
\begin{center}
\begin{tabular}{ccccc}
Method & Pose & IR & CS & CV \\
\hline \hline
Pose network & X & - & 82.3 & 89.5 \\ 
\end{tabular} 
\end{center}
\end{table}

\begin{table}[t]
\caption{Results of the IR module on NTU RGB+D dataset  (accuracy in \%)}
\label{ir_block_results}
\begin{center}
\begin{tabular}{ccccc}
Method & Pose & IR & CS & CV \\
\hline \hline
IR network & - & X & 89.8 & 94.1 \\ 
\end{tabular} 
\end{center}
\end{table}

For consistency, we do not modify the following hyperparameters across all experiments. We set the batch size to 16 which allows the model and a batch to fit on most high-end GPUs. Gradient clipping is used to avoid an exploding gradient issue. We set it to 10. Adam optimizer \cite{kingma2014adam} is used to train the networks. A learning rate of 0.0001 is set and kept consistent during training. 

The pose and IR modules each require a fixed input size. Skeleton maps are resized to 224x224 images. IR frames are resized to 112x112. 

To assure consistency and reproducibility, we use a pseudorandom number generator fed with a fixed seed. Following \cite{shahroudy2016ntu}, we sample 5\% of the training set as our validation set.  

\begin{table}[t]
\caption{Impact of pre-training on classification performances (A: Augmented | P: Pre-trained | C: cropped inputs) (accuracy in \%)}
\label{pretraining_results}
\begin{center}
\begin{tabular}{ccccc}
Method & Pose & IR & CS & CV \\
\hline \hline
Pose module & X & - & 78.7 & 85.1 \\
Pose module - P & X & - & \textbf{80.7} & \textbf{87.0} \\
\hline
IR module & - & X & 76.8 & 76.3 \\
IR module - P & - & X & \textbf{84.0} & \textbf{84.6} \\
\hline
IR module - C & - & X & 84.6 & 88.6 \\
IR module - CP & - & X & \textbf{90.1} & \textbf{91.2} \\ 
\end{tabular} 
\end{center}
\end{table}

\begin{table}[t]
\caption{Impact of data augmentation on classification performances (A: Augmented | P: Pre-trained | C: cropped inputs) (accuracy in \%)}
\label{augmentation_results}
\begin{center}
\begin{tabular}{ccccc}
Method & Pose & IR & CS & CV \\
\hline \hline
Pose module - P & X & - & 80.7 & 87.0 \\
Pose module - PA & X & - & \textbf{82.3} & \textbf{89.5} \\
 \hline
IR module - P & - & X & 84.0 & 84.6 \\
IR module - PA & - & X & \textbf{85.0} & \textbf{87.5} \\
 \hline
IR module CP & - & X & \textbf{90.1} & 91.2 \\
IR module CPA & - & X & 89.8 & \textbf{94.1} \\
 \hline
FUSION - CP & X & X & 90.8 & 94.0 \\
FUSION - CPA & X & X & \textbf{91.6} & \textbf{94.5} \\ 
\end{tabular}
\end{center}
\end{table}

\begin{table}[t]
\caption{Impact of our cropping strategy on classification performances (A: Augmented | P: Pre-trained | C: cropped inputs) (accuracy in \%)}
\label{cropping_results}
\begin{center}
\begin{tabular}{ccccc}
Method & Pose & IR & CS & CV \\
\hline \hline
IR module & - & X & 76.8 & 76.3 \\
IR module - C & - & X & \textbf{84.6} & \textbf{88.6} \\
 \hline
IR module- PA & - & X & 85.0 & 87.5 \\
IR module - CPA & - & X & \textbf{89.8} & \textbf{94.1} \\ 
\end{tabular}
\end{center}
\end{table}

\subsection{Ablation studies}

In this section, we isolate the pose and IR modules and study their individual contribution with regard to different parameters. Action classification accuracy on the NTU RGB+D dataset is used as the comparison metric. We evaluate the impact of transfer learning, data augmentation, pose conditioning of IR sequences and the number of frames $T$. Finally, we compare our results with the current state of the art.

\begin{table*}[t]
\caption{Impact of IR sequence length on classification performances (A: Augmented | P: Pre-trained | C: cropped inputs) (accuracy in \%)}
\label{seq_len_results}
\begin{center}
\begin{tabular}{ccc|cccc|cccc}
& & & \multicolumn{4}{c|}{ CS} & \multicolumn{4}{c}{ CV } \\
\cline{4-11}
Method & Pose & IR & T=8 & T=12 & T=16 & T=20 & T=8 & T=12 & T=16 & T=20 \\ 
\hline \hline
IR module - CPA & - & X & 86.8 & 89.5 & 90.0 & 89.8 & 88.8 & 91.3 & 93.0 & 94.1 \\
FUSION - CPA & X & X & \textbf{88.7} & \textbf{90.4} & \textbf{90.3} & \textbf{91.6} & \textbf{92.4} & \textbf{94.4} & \textbf{94.3} & \textbf{94.5} \\ 
\end{tabular} 
\end{center}
\end{table*}

\subsubsection{Pose module}

We evaluate the performances of our pose module as a standalone. The IR module does not intervene. We also adjust the input size of the classification MLP. Optimal results are achieved by combining pre-training with data augmentation. Table \ref{skeleton_block_results} shows the best results of the pose module on NTU RGB+D: 82.3\% on CS and 89.5\% on CV. 

The CV benchmark is a much easier task, hence the better results compared to CS. The test actions are already seen during training but from a different point of view with a different camera. Although the different setups yield different joint position estimations for a given sequence \cite{zhang2019view}, the geometric nature of skeleton data allows for a better generalization. This is not the case for the CS task as the test sequences are completely novel. Consequently, the following discussions will only address the CS benchmark.

The confusion matrix reveals the pose module's strong ability to correctly classify actions with intense kinetic movements. Actions such as sitting down, standing up, falling, jumping, staggering, walking toward or away from another subject are classified with over 95\% accuracy. Unsurprisingly, actions with similar skeleton motions prove the most challenging. Writing is the trickiest, with 40\% accuracy only and often mislabeled as writing or typing on a keyboard. The incorrectly classified actions fall under two categories: similar motion actions and object-related actions. We believe this will always be a limitation of pose-only networks.

\subsubsection{Infrared module}

The other part of the FUSION network, and arguably the most important contributor, is the infrared module. In similar fashion as above, the input size of the MLP is adjusted while keeping the number of neurons equal. Optimal results are achieved with a pre-trained network, with data augmentation, on pose-conditioned inputs for a sequence length of $T=20$. Table \ref{ir_block_results} shows the performance of the IR module as a standalone: 89.8\% on CS and 94.1\% on CV. 

The confusion matrix reveals a more balanced accuracy score over the different actions of the NTU RGB+D dataset. Some actions, such as touching another person's pocket or staggering, prove more difficult to recognize for the IR module compared to the pose module. This reinforces our intuition that pose and visual streams are complementary. However, some object-oriented actions are still difficult to correctly discern. For instance, writing is more often than not mislabeled as playing with a phone. We propose two possible explanations. Firstly, the object information might be lost during the rescaling process, even with our cropping strategy in place. Secondly, the IR nature, grayscale and noisy, might not be clear enough to discern the object correctly. But other object-related actions such as dropping an object or brushing hair see an impressive improvement of over 10\%. 

\subsubsection{Influence of pre-training}

Pre-training a network is an elegant way to transfer a learned task to a new one. It has been shown to provide impressive results even on handcrafted images \cite{zhang2019view}. Furthermore, it helps with the overfitting issue smaller datasets may demonstrate.

We evaluate the impact of this strategy on our network. Table \ref{pretraining_results} shows the effect of pre-training on the different modules. 

The pose network enjoys a noticeable increase in accuracy of about 2\% for both benchmarks (78.7\% to 80.7\% on CS). It is pre-trained on ImageNet, which consists of real-life images. The skeleton maps used as inputs are handcrafted. Even then, a pre-training scheme shows encouraging results.

The impact of pre-training on the IR module's accuracy is significant. For uncropped sequences, the accuracy increases by about 7\% for both benchmarks (76.8\% to 84.0\% on CS). For cropped sequences, the gain is over 5\% for the cross-subject benchmark (84.6\% to 90.1\%) and almost 3\% for cross-view (88.6\% to 91.2\%). 

The greater contribution of transfer learning for the IR module compared to the pose module might be explained by the greater resemblance of IR vs. RGB videos compared to handcrafted vs. real-life images. Nonetheless, such findings further emphasize the power of transfer learning. 

\subsubsection{Influence of data augmentation}

Data augmentation consists of virtually enlarging the dataset, thus hopefully preventing overfitting and reducing variance between training and test sets. We perform augmentation for the different data streams. Table \ref{augmentation_results} shows the performances of data augmentation on the different modules with pre-trained networks. Overall, data augmentation yields favorable results.

The pose module alone enjoys an increase of about 2\% accuracy for both benchmarks (80.7\% to 82.3\% on CS). The IR module alone seems to benefit more from data augmentation on the CV benchmark compared to the CS. For the CV benchmark, the increase is about 3\% whether the input sequence is cropped (91.2\% to 94.1\%) or not (84.6\% to 87.5\%). For the CS benchmark, the improvements are not significant. When the modules are fused, our FUSION network, data augmentation is favorable but not significant with an increase in the 0.5\% range (90.8\% to 91.6\% on CS). However, this could be expected. As the baseline results increase, the gains are expected to diminish. 

\subsubsection{Transfer learning vs. data augmentation}

Transfer learning and data augmentation are two strategies to better generalize the performances of a network. Transfer learning leverages the learned parameters from another dataset while data augmentation virtually enlarges the current dataset. A small dataset might lead to overfitting which results in an increase in variance between the training and validation sets as the training error continues to lower. 

Our model is able to reach a negligible training error, even with individual modules, showcasing an overfitting issue. Having studied the impacts on performances of both methods, transfer learning shows much better results. This might be explained by the already large size of the NTU RGB+D dataset mitigating the potential of data augmentation. Nonetheless, it is formidable how a model can yield vastly different performances based on the initialization of its parameters. The black box nature of deep learning makes the interpretation of how a model learns difficult. Perhaps future works will focus on understanding the internal representation of a network to guide its learning rather than implementing evermore complex models. 

\subsubsection{Influence of pose-conditioned cropped IR sequences}

In this section, we evaluate the impact of our cropping strategy, detailed section \ref{cropping_strategy}, on the performances of the IR module as a standalone. Table \ref{cropping_results} shows a significant increase in performances. 

Our baseline for this comparison, the IR module without transfer learning and data augmentation on uncropped sequences, reports unsatisfactory results (76.8\% on CS). With transfer learning and data augmentation, we are able to increase the accuracy by almost 10\% average for both benchmarks (76.8\% to 85.0\% on CS). However, we find that our cropping strategy alone reaps similar benefits (76.8\% to 84.6\% on CS). When combining all three strategies, we further ameliorate the classification score by about 5\% (89.8\% on CS). The average gain for both benchmarks is thus above 15\%, which is considerable.

We demonstrate the power of a pragmatic approach. An identical model is able to perform significantly better thanks to careful design choices.

\subsubsection{Influence of sequence length}

Sequences of the NTU RGB+D dataset are at most a couple of seconds long. We study the impact of the length $T$ of the new sampled IR sequence on classification performances of two networks: the IR module only and on the complete FUSION model. Both models are pre-trained and fed with augmented data. The IR sequences are pose-conditioned. Table \ref{seq_len_results} reports the impact of different values of $T$ on the accuracy score. 

As a general tendency, the greater the value of $T$, the better the results. Best results are achieved for $T=20$ for three out of four scenarios (on CS: 89.8\% for IR module only and 91.6\% for FUSION). The exception happens for the IR module as a standalone on the CS benchmark where the optimal value is $T=16$ (90.0\%). However, the difference in accuracy is negligible. For the FUSION network, excellent results are achieved for a number of frames as little as $T=12$ (90.4\% on CS and 94.4\% on CV). FUSION networks with a smaller value of $T$ are much faster, showcasing a trade-off between speed and accuracy. 

\subsubsection{Comparison with the state of the art}
\begin{table}[t]
\caption{Comparison of our model to the state of the art (A: Augmented | P: Pre-trained | C: cropped inputs) (accuracy in \%)}
\label{state_of_the_art}
\begin{center}
\begin{tabular}{ccccccc}
Method & Pose & RGB & Depth & IR & CS & CV \\
\hline \hline
Lie Group \cite{vemulapalli2014human} & X & - & - & - & 50.1 & 82.8 \\
\hline
HBRNN \cite{du2015hierarchical} & X & - & - & - & 59.1 & 64 \\
Deep LSTM \cite{shahroudy2016ntu} & X & - & - & - & 60.7 & 67.3 \\
PA-LSTM \cite{shahroudy2016ntu} & X & - & - & - & 62.9 & 70.3 \\
ST-LSTM \cite{liu2016spatio} & X & - & - & - & 69.2 & 77.7 \\
STA-LSTM \cite{song2017end} & X & - & - & - & 73.4 & 81.2 \\
VA-LSTM \cite{zhang2017view} & X & - & - & - & 79.2 & 87.7 \\
\hline
TCN \cite{kim2017interpretable} & X & - & - & - & 74.3 & 83.1 \\
C+CNN+MTLN \cite{ke2017new} & X & - & - & - & 79.6 & 84.8 \\
Synthesized CNN \cite{liu2017enhanced} & X & - & - & - & 80 & 87.2 \\
3scale ResNet \cite{li2017skeleton} & X & - & - & - & 85 & 92.3 \\
\hline
DSSCA-SSLM \cite{shahroudy2017deep} & - & X & X & - & 74.9 & - \\
\cite{rahmani2017learning} & X & - & X & - & 75.2 & 83.1 \\
CMSN \cite{zolfaghari2017chained} & X & X & - & - & 80.8 & - \\
STA-HANDS \cite{baradel2017pose} & X & X & - & - & 84.8 & 90.6 \\
Coop CNN \cite{wang2018cooperative} & - & X & X & - & 86.4 & 89 \\
\hline
ST-GCN \cite{yan2018spatial} & X & - & - & - & 81.5 & 88.3 \\
DGNN \cite{shi2019skeleton} & X & - & - & - & 89.9 & \textbf{96.1} \\
\hline
\textbf{Pose module - PA} & X & - & - & - & 82.3 & 89.5 \\
\textbf{IR module - CPA} & - & - & - & X & 89.8 & 94.1 \\
\textbf{FUSION - CPA} & X & - & - & X & \textbf{91.6} & 94.5 \\ 
\end{tabular} 
\end{center}
\end{table}

We compare our FUSION model with the state of the art (Table \ref{state_of_the_art}). We divide current methods into 5 different frameworks including handcrafted features, RNN-based methods, CNN-based methods, fusion methods and GCN-based methods. Current best results are obtained using skeleton data only with GCNs. We achieve better results than the current state of the art on the CS benchmark (91.6\%) with 1.7\% accuracy increase. On the CV benchmark, results are comparable (94.5\% for FUSION against 96.1\% for DGNN \cite{shi2019skeleton}). We conclude to the efficacy of IR data to correctly interpret human actions. 

We significantly improve upon current fusion methods, once again validating the complementary role of pose and visual data. 

\section{Conclusion}

We propose an end-to-end trainable network using skeleton and infrared data for human action recognition. A pose module extracts features from skeleton data and an infrared module learns from videos. The 3D skeleton is used as an input source and also conditions the infrared stream, providing a crop around the subjects. The two stream features are then concatenated, and a final prediction is outputted. The pose and infrared modules report strong individual performances, which is greatly due to the power of transfer learning as they are both pre-trained on other large scale datasets. When working in symbiosis, the results are further ameliorated. We are the first to conjointly use pose and infrared streams. Our method improves the state of the art on the largest RGB-D action recognition dataset to date. Compared to other stream fusion approaches, our method requires less prepossessing and is more memory efficient. 

Our work demonstrates the strong representational power of infrared data, which opens the door for applications where illumination conditions render RGB videos unusable. The complementary role of pose and visual streams is further illustrated, which is in line with previous work. Given the modular nature of our proposed network, future works could focus on more modern pose modules such graph neural networks.

\section*{Acknowledgment}

This work was supported by research grants from the Natural Sciences and Engineering Research Council of Canada and an industrial funding from Aerosystems International Inc. The authors would also like to thank their collaborators from Aerosystems International Inc.





\bibliography{bibliography}
\bibliographystyle{plain}

\end{document}